\documentclass[letterpaper]{article} 
\usepackage{aaai2027}  
\nocopyright
\usepackage[hyphens]{url}  
\usepackage{graphicx} 
\usepackage{natbib}  
\usepackage{caption} 
\usepackage{amsmath}
\usepackage{amssymb}
\usepackage{booktabs}

\title{Escaping the Phylogenetic Trap: Persistent Homology Enables Cross-Species Ion Channel Classification via Multimodal Graph Neural Networks}

\author{
    Arezoo Bybordi\textsuperscript{\rm 1},
    Favour Achimba\textsuperscript{\rm 1},
    Mariam Gelashvili\textsuperscript{\rm 2},
    Jessy Ramirez\textsuperscript{\rm 2,\rm 3},
    Weigang Qiu\textsuperscript{\rm 2,\rm 4},
    Mand\"{e}~Holford\textsuperscript{\rm 1,\rm 5,\rm 6},
    Anita Raja\textsuperscript{\rm 2}
}
\affiliations{
    \textsuperscript{\rm 1}The Graduate Center, City University of New York\\
    \textsuperscript{\rm 2}Hunter College, City University of New York\\
    \textsuperscript{\rm 3}Tandon School of Engineering, New York University\\
    \textsuperscript{\rm 4}Weill Cornell Medicine\\
    \textsuperscript{\rm 5}Harvard University\\
    \textsuperscript{\rm 6}American Museum of Natural History\\
    arezoo.bybordi26@login.cuny.edu
}
\begin{document}
\maketitle

\begin{abstract}
Protein classifiers are highly sensitive to evaluation protocol: random train/test splits induce phylogenetic leakage, allowing models to exploit taxon-specific correlations and substantially overestimate out-of-distribution performance. This issue is particularly acute for venom compound classification by ion-channel target (Ca$^{2+}$, K$^+$, Na$^+$), where reported accuracies of 80–90\% do not reflect generalization to unseen taxa. We formulate cross-species prediction as a structured distribution shift in which taxonomic identity acts as a confounder, requiring representations that capture structural features invariant across phylogeny. Using a shared multimodal backbone that fuses ESM-2 sequence embeddings with a disulfide-aware graph attention network, we systematically compare five structural inductive biases—spectral, Riemannian, SE(3)-invariant, statistical-manifold, and persistent-homology representations—under a leave-one-taxon-out evaluation protocol on a benchmark of 1,373 ColabFold-predicted venom compounds spanning five taxa. The persistent-homology representation summarizes backbone geometry through topological descriptors that are robust to local structural perturbations, providing an inductive bias toward phylogenetically invariant learning. Under conventional i.i.d. splits, all models achieve similar accuracy (81–86\%), masking differences in inductive bias. Under taxonomic shift, this parity collapses: a sequence-only baseline drops to approximately 30\%, the remaining geometric priors cluster at 54–55\%, while our persistent-homology model (PH-GNN) achieves 63.4\%, outperforming the next-best method by over 8 percentage points. These results demonstrate that random-split evaluation obscures meaningful differences in representation quality and identify topological representations as an effective inductive bias for learning under structured biological distribution shift.

\end{abstract}

\section{Introduction}

Ion channels are critically important therapeutic targets for channelopathies, playing foundational roles in the pathophysiology of epilepsy, cancer, and pain management. Consequently, identifying which proteins or peptides target specific ion channels is a central objective in drug discovery. Venomous organisms have independently evolved ion channel-targeting compounds across at least a dozen phyla over hundreds of millions of years \cite{Casewell2013,King2011}. This convergent evolution produces a striking biological paradox: calcium, potassium, and sodium channel compounds from cone snails, spiders, scorpions, sea anemones, and snakes accomplish related functional disruptions through wildly divergent primary sequences and 3D folds \cite{Olivera2007, Klint2012}.
For AI-driven drug discovery, capturing this structural and sequential flexibility requires robust representation learning paradigms. Specifically, these paradigms must learn to bypass surface-level sequence homology to identify the invariant topological features that actually dictate target binding. Previous computational efforts to map protein-ion channel interactions have utilized Variational Autoencoders (VAEs) \cite{vae}, physicochemical property mapping \cite{chou2001prediction}, sequence-based Evolutionary Scale Modeling (ESM) \cite{Lin2023} + Random Forest architectures \cite{Achimba2025}, and family-specific tools like ICTCPred \cite{Zhang2016}. However, these deep learning models share a critical vulnerability: an inability to generalize to unseen proteins. Because they are typically evaluated under random train/test splits, classifiers tend to learn phylogenetic features rather than functional ones. We define the \textit{phylogenetic trap} as the phenomenon where models achieve artificially inflated accuracy by memorizing evolutionary lineage leaked across random train/test splits, rather than learning generalized functional features. In practice (such as screening newly sequenced venom transcriptomes from uncharacterized species), a method that achieves 85\% accuracy under a random split but 6\% on an unseen taxon provides no scientific value. 

We address this gap by asking a fundamental research question: \textit{What mathematical representations of protein structure best capture biological evolution and functional generalization?} To answer this, we propose a novel multimodal framework that bridges mathematics and biology. To our knowledge, we are the first to fuse Evolutionary Scale Modeling (ESM-2) sequence embeddings with a Graph Attention Network (GAT) that explicitly encodes disulfide bond weights, utilizing ColabFold-predicted structures to predict $Na^{+}$, $K^{+}$, and $Ca^{2+}$ channel targets.

Our central argument is that the choice of geometric inductive bias determines a model's susceptibility to the phylogenetic trap. Geometric features, spectral modes, and sequence distributions are inherently shaped by the evolutionary trajectory of the organism, leading to severe taxa-entanglement. By contrast, the three-dimensional topology of a compound's interaction surface is constrained strictly by the deeply conserved geometry of the target channel's pore \cite{Doyle1998, MacKinnon2003}. 

\textbf{We hypothesize that persistent homology, acting as a higher-order abstraction that captures multi-scale cavity topology, encodes this evolutionarily invariant signature, enabling true zero-shot cross-species generalization.} In this work, our core contributions are:
\begin{enumerate}

\item \textbf{A novel multimodal architecture:} We introduce a unified GNN framework fusing ESM-2 embeddings with structure-aware GATs and explicit disulfide bond attention, evaluated on a curated dataset of 1,373 ColabFold-predicted compounds across five diverse taxa.

\item \textbf{A systematic evaluation of geometric priors:} We ablate our baseline against five distinct mathematical frameworks: statistical manifolds (Fisher information), persistent homology, spectral graph theory (Laplacian eigenvectors), Riemannian manifolds, and $SE(3)$ invariance.

\item \textbf{Escaping the Out-Of-Distribution (OOD) trap:} We demonstrate that while statistical manifolds achieve state-of-the-art accuracy (86.2\%) on random splits, they collapse out-of-distribution. Conversely, our persistent homology framework achieves a leading 63.4\% cross-species accuracy, proving that topology generalizes where local geometry fails.
\end{enumerate}
We call our complete framework PH-GNN (Persistent Homology Graph Neural Network). Code and data are available at \url{https://github.com/deleteblue/PH_GNN}.
\section{Related Work}

The most direct prior approach to our task is the recent Molecular Arms Race Classifier (MARC) of \citet{Achimba2025}, which predicts sodium, potassium, and calcium ion channel targets from cysteine-rich venom peptides using mean-pooled ESM-2 features fed into a Random Forest. Earlier family-specific tools have similarly leveraged sequence-derived features and classical classifiers for narrower scopes, such as ICTCPred, which combines SMOTE oversampling with hybrid sequence features for conotoxin ion-channel-target prediction \cite{Zhang2016}. All of these methods are evaluated under random train-test splits, which permit homologous proteins from the same taxon to leak between training and test partitions. Our work treats this evaluation regime as the central object of study rather than as background. Self-supervised protein language models such as ESM-2 \cite{Lin2023} produce sequence embeddings that have become the de facto input feature for downstream prediction tasks. Recent structure-aware variants integrate predicted-structure tokens into the vocabulary \cite{Su2024}, and graph-based encoders such as the Geometric Vector Perceptron \cite{Jing2021} learn over 3D atomic neighborhoods with $SE(3)$-equivariant message passing. Our multimodal backbone composes ESM-2 with a disulfide-aware GAT \cite{Velickovic2018} in this established tradition; the architectural contribution lies not in any single component but in the controlled comparison of structural inductive biases that share a common backbone.

Topological data analysis has a substantial history in molecular property prediction, evolving from early protein-ligand models \citet{Cang2017} to recent graph-based applications (\cite{Arafat_Basu_Gel_Chen_2025}, \cite{Deng_Ma_Zhang_Jia_Bing_2026}). However, most prior work maximizes in-distribution accuracy using high-dimensional descriptors rather than testing cross-distribution robustness.

We deliberately retain a low-dimensional summary-statistic representation, motivated by the hypothesis that coarse topological invariants of backbone geometry carry the cross-taxon signal while richer descriptors risk re-encoding lineage-specific structure. The general problem of inflated random-split accuracy under distribution shift has been documented across domains \cite{Koh2021}, and protein-specific benchmarks such as FLIP \cite{Dallago2021} have established sequence-identity and engineering-relevant splits as standard for fitness prediction. Our taxa leave-one-out protocol applies this evaluation philosophy to the domain of cross-species compound classification, where phylogeny rather than sequence identity is the dominant axis of distributional shift relevant to deployment.

\section{Problem Formulation}
Let $\mathcal{D} = \{(G_i, e_i, y_i, t_i)\}_{i=1}^N$ be a dataset where $G_i$ is the protein structure graph, $e_i \in \mathbb{R}^{1280}$ is the ESM-2 embedding, $y_i \in \{\text{Calcium}, \text{Potassium}, \text{Sodium}\}$ is the ion channel target class, and $t_i \in \mathcal{T} = \{\text{Snake}, \text{Cone Snail}, \text{Spider}, \text{Scorpion}, \text{Sea Anemone}\}$ is the organism taxon. Random split evaluation draws train/test partitions uniformly at random (80/20, stratified by class). This allows proteins from the same species and potentially the same protein family to appear in both splits. Taxa OOD evaluation holds out all proteins from a single taxon $t^* \in \mathcal{T}$:
$\mathcal{D}_{\text{train}} = \{i : t_i \neq t^*\}, \quad \mathcal{D}_{\text{test}} = \{i : t_i = t^*\}.$
We perform five-fold leave-one-taxon-out cross-validation and report mean accuracy and standard deviation. This measures the ability to classify compounds from an organism type never seen during training.
\section{Dataset}

We curated our primary dataset by sampling from the foundational database established by \citet{Achimba2025}. To establish a focused and rigorous classification benchmark, we restricted our scope to three primary ion channel targets ($Na^{+}$, $K^{+}$, and $Ca^{2+}$), yielding 1,376 proteins. Structures were predicted using ColabFold. ColabFold is a tool that predicts protein structures using AlphaFold. AlphaFold is a deep learning system developed by Google DeepMind that predicts the 3D atomic coordinates of a protein directly from its primary amino acid sequence with near-experimental accuracy \cite{Jumper2021}. Three proteins were excluded due to cross-dataset label conflicts (identical sequences assigned to different ion channel classes), resulting in a final dataset of $N = 1,373$ unique compounds. The dataset exhibits a natural class imbalance: Calcium ($n=216$, 15.7\%), Potassium ($n=542$, 39.5\%), and Sodium ($n=615$, 44.8\%). The $\sim$3:1 imbalance between the calcium minority class and the majority classes is explicitly addressed via a class-weighted cross-entropy loss during training. For our taxa leave-one-out out-of-distribution (OOD) evaluation, the approximate test-fold sizes are distributed across five diverse taxa: Snake ($n=77$), Cone Snail ($n=127$), Sea Anemone ($n=161$), Spider ($n=362$), and Scorpion ($n=646$). This variation reflects the natural unevenness of venom databases, making our OOD evaluation highly representative of real-world deployment. All 3D structures were predicted using ColabFold \cite{Mirdita2022} powered by AlphaFold2 \cite{Jumper2021} with MMseqs2 multiple sequence alignments \cite{steinegger2017mmseqs2}. We utilized the top-ranked unrelaxed predicted structure for each protein. To preserve full dataset coverage across understudied taxa, structure quality was not filtered by Predicted Local Distance Difference Test (pLDDT) \cite{Jumper2021} confidence scores. While this introduces structural noise, it accurately reflects real-world conditions where high-resolution experimental structures are unavailable for novel species. Atom-level graphs are constructed as radius graphs, connecting any two atoms within $4.0$~\AA{}; this captures all covalent bonds as well as close non-bonded contacts such as hydrogen bonds. Crucially, we explicitly encode disulfide bonds as a distinct edge type, detected via an S-S distance threshold of $2.5$ \AA. Disulfide bonds are functionally critical to the thermodynamic stability of compound folds. Node features are 8-dimensional, comprising a 5-dimensional one-hot encoding of atom types (C, N, O, S, other) and 3-dimensional Cartesian coordinates normalized per-protein.
\section{PH-GNN Architecture}

To isolate the impact of different structural representations on cross-species generalization, all six evaluated architectures share a common sequence encoder and fusion-classifier block. They differ exclusively in their structural encoding strategy. 

\begin{figure*}[t]
    \centering
    \includegraphics[width=\textwidth]{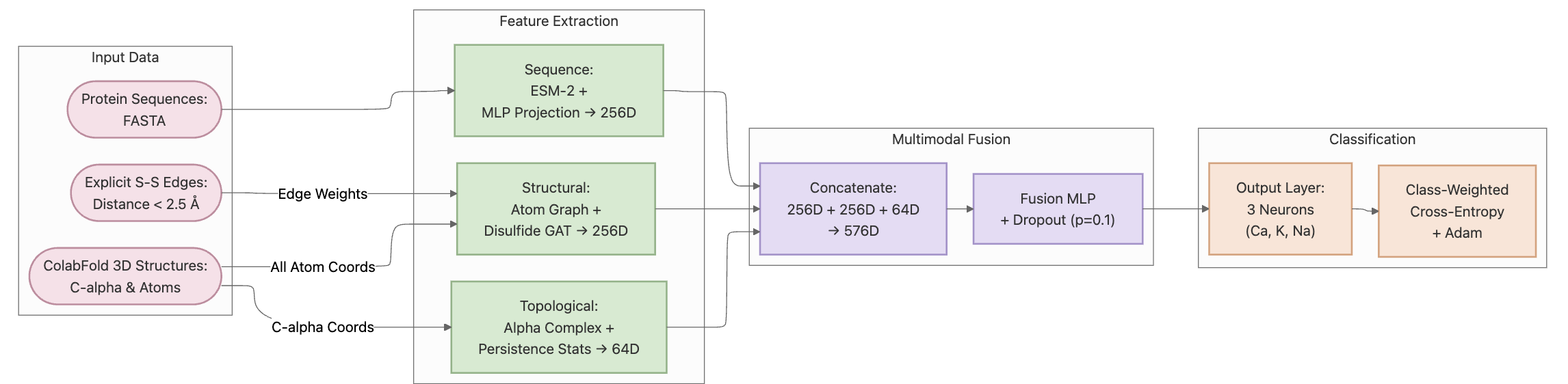}
    \caption{\textbf{PH-GNN Multimodal Architecture (M2).} Our framework extracts three parallel representations from the ColabFold-predicted input data: (1) sequence embeddings via a pre-trained ESM-2 language model. The three modalities are concatenated into a joint representation and passed through a fusion Multi Layer Perceptron (MLP) to predict the target ion channel class ($Ca^{2+}$, $K^{+}$, $Na^{+}$), (2) structural embeddings via a disulfide-aware Graph Attention Network (GAT) operating on all-atom proximity graphs, and (3) topological features via Alpha complex filtration of $C\alpha$ coordinates. }
    \label{fig:architecture}
\end{figure*}

\subsection{Shared multimodal backbone}
\textbf{(1) Sequence encoder.} We utilize Evolutionary Scale Modeling (ESM-2, 650M parameters) \cite{Lin2023} to extract sequence features. Mean-pooled embeddings ($e_i \in \mathbb{R}^{1280}$) are projected to $d=256$ dimensions via a two-layer MLP with LayerNorm and ReLU activations. ESM-2 weights remain frozen during training. We use ESM because as a protein language model, it inherently encodes the evolutionary trajectory and mutational constraints of the primary sequence. \textbf{(2) Baseline structural encoder.} The reference structural encoder is a two-layer Graph Attention Network (GAT) \cite{Velickovic2018} with 4 attention heads. Crucially, the GAT explicitly processes disulfide bonds via learnable edge embeddings added to the message-passing step, capturing fold stability. Global mean-pooling produces the structural representation $z_{struct} \in \mathbb{R}^{256}$. \textbf{(3) Fusion and classifier.} The sequence representation $z_{seq}$ and structural representation $z_{struct}$ are concatenated and passed through a two-layer MLP with dropout ($p=0.1$) to produce a 256-dimensional joint embedding, which is linearly projected to the three target logits.

\subsection{(M1) The Phylogenetic Foil: Statistical Manifold Encoder}
To maximize discriminative power within the training distribution, we treat the ESM-2 embedding space as a parametric statistical manifold. This encoder modifies both the sequence branch and the fusion attention mechanism to respect the geometry of the Fisher information manifold. The ESM-2 projection layer is trained using the natural gradient \cite{Amari1998} rather than vanilla gradient descent. The natural gradient is preconditioned by the Fisher information matrix $F(\theta) = \mathbb{E}[\nabla \log p(x|\theta)\nabla \log p(x|\theta)^\top]$, approximated per mini-batch using the Kronecker-factored approximation \cite{Martens2015}. This optimization is geometry-aware: updates traverse equal statistical distances rather than Euclidean distances, accelerating convergence in directions of low Fisher curvature. In the fusion module, the standard dot-product attention is replaced by a Fisher-metric inner product \cite{Amari2016}:
\begin{equation}
\text{attn}(z_{seq}, z_{struct}) = z_{seq}^\top F^{-1/2} z_{struct}
\end{equation}
where $F^{-1/2}$ is the inverse square root of the sequence encoder's Fisher information matrix. 

\textbf{Theoretical intuition:} This architecture weights attention by statistical confidence, amplifying directions of high Fisher curvature that are well-constrained by the training data. While this theoretically provides optimal discriminative power for in-distribution high-throughput screening, we hypothesize that it creates a ``phylogenetic trap.'' Because the Fisher matrix is derived purely from sequence distributions, the model risks extrapolating blindly in Fisher space when presented with an out-of-distribution (OOD) taxon, potentially leading to catastrophic failure.

\subsection{(M2) Escaping the trap: PH-GNN with persistent homology encoder}
To construct a representation robust to evolutionary distributional shifts, we augment the structural branch with a topological descriptor that is entirely amino-acid agnostic.

For each protein, we extract C$\alpha$ atom coordinates from the ColabFold structure and compute an alpha complex filtration using GUDHI \cite{TheGUDHIProject2015}, truncated at a squared circumradius of $\alpha^{2} = 400$~\AA$^{2}$ (circumradius $20$~\AA), discarding features with persistence below $0.1$~\AA$^{2}$. The resulting persistence diagram tracks the birth and death of topological features (connected components, loops, and voids) as the filtration scale grows. We summarize the diagram into a 12-dimensional vector by computing four lifespan statistics for each of the three Betti channels ($H_0$ connected components, $H_1$ loops, $H_2$ voids): the feature count, the maximum lifespan, the mean lifespan, and the total persistence. Features with infinite death time are assigned a fixed lifespan of 50. This topology vector $p_i \in \mathbb{R}^{12}$ is projected to 64 dimensions via a linear layer and concatenated with the GAT output prior to fusion.
\textbf{Theoretical intuition:} Persistent homology measures the invariant shape of the backbone loop structure at multiple scales. We argue that the topology of the interaction surface, specifically how loops arrange to engage the ion channel pore, is constrained by the deeply conserved geometry of the pore itself \cite{Doyle1998, MacKinnon2003}. By operating purely on continuous multi-scale geometry rather than sequence distributions, persistent homology theoretically encodes a convergent structural signature. This provides the necessary inductive bias to recognize functional similarity across entirely different evolutionary lineages, bypassing the phylogenetic trap.
\subsection{Alternative Geometric Priors}
To ensure a comprehensive evaluation of structural inductive biases, we ablate against three additional graph architectures: \textbf{(1) Spectral (M3):} Replaces the GAT with spectral convolutions using the $k=32$ lowest eigenvectors of the normalized graph Laplacian \cite{Defferrard2016}, encoding global structural modes. \textbf{(2) Riemannian (M4):} Learns a protein-specific Riemannian metric tensor to reweight message aggregation, replacing Euclidean distances with geodesic distances in the attention kernel. \textbf{(3) Invariant $SE(3)$ (M5):} Replaces Cartesian node features with provably $SE(3)$-invariant descriptors (inter-atomic distances, $C\alpha$ bond angles, and backbone dihedrals) \cite{Jing2021}, forcing the network to focus on relative geometry rather than absolute spatial orientation.

\section{Experiments}
To systematically evaluate the susceptibility of different structural representations to the phylogenetic trap, we subject all architectures to identical training and evaluation pipelines. 

\textbf{Evaluation Protocols.} 
We contrast model performance under two distinct regimes. For the \textit{in-distribution} evaluation, we utilize a standard 80/20 random train/test split. To explicitly test cross-species generalization (\textit{out-of-distribution}), we employ a rigorous taxa leave-one-out protocol (5 folds). For each OOD fold, a separate model is trained from scratch on four taxa and evaluated exclusively on the held-out fifth taxon. Absolutely no information from the test taxon is available during training. All evaluated models are subjected to identical data splits; optimizer settings are shared except where noted in Training Details.
We report overall accuracy, per-class F1, and macro-averaged F1. Macro F1 serves as our primary metric to account for the class imbalance (Calcium: 216, Potassium: 542, Sodium: 615) and to explicitly expose minority-class collapse under distribution shift. To ensure robust conclusions, statistical significance is rigorously established using exact McNemar tests on pooled per-protein predictions, with Holm-Bonferroni correction applied across all baseline comparisons.

\textbf{Baselines.} 
We compare our five geometric and topological ESM-GNN fusion variants against the standard ESM+GAT baseline, where a Graph Attention Network explicitly processes disulfide bonds. Additionally, to quantify the absolute necessity of structural modalities, we implement a strong sequence-only baseline following the exact methodology of \citet{Achimba2025}: an ESM + Random Forest (RF) classifier. This model classifies the identical mean-pooled ESM-2 features using a Random Forest (94 trees, max features = $\sqrt{d}$) trained on the exact same data splits.

\textbf{Training Details.} 
All neural architectures are trained using a class-weighted cross-entropy loss to mitigate the natural 3:1 imbalance of the calcium target class. We optimize the networks using Adam with a learning rate of $\eta = 10^{-3}$ and dropout $p = 0.1$. The one exception is the ESM+GAT baseline under OOD evaluation, which was trained with additional regularization ($\eta = 3 \times 10^{-4}$, weight decay $10^{-4}$, dropout $p = 0.15$) to reduce memorization of the training taxa; because this configuration favors the baseline under distribution shift, the reported OOD comparison is conservative with respect to PH-GNN. Training is capped at a maximum of 50 epochs and governed by early stopping based on validation loss with a patience of 10 epochs. To follow the baseline from \citet{Achimba2025}, during the random-split evaluation, SMOTE \cite{Chawla2002} oversampling is applied strictly to the training set. Architecture dimensions and batch sizes were fixed a priori. We performed a minimal grid search over learning rate, weight decay, and dropout, selecting the final configuration based on mean OOD validation accuracy.  Random Forest baseline underwent a grid search from 10 to 200 estimators.

\textbf{Reproducibility.} 

Reported results reflect a single training run per model. Data splits are deterministically seeded (random\_state=42); however, network initializations are unseeded. Variance is rigorously quantified via exact McNemar tests and bootstrap resampling (10,000 resamples; 95\% CIs) over fixed test sets. Experiments ran on PSC Bridges-2 CPU nodes (8 cores, 15GB RAM, Ubuntu 22.04, PyTorch 1.13.1). Training requires no GPUs, completing a full 5-fold sweep in approximately 3.3 CPU-hours per model.
\section{Results}

\subsection{In-Distribution: Random split performance}
Table \ref{tab:random_split} reports test performance under the standard random 80/20 split. All multimodal GNN architectures substantially outperform the sequence-only ESM+RF baseline and occupy a narrow 81 to 86\% accuracy band. The \textbf{Statistical Manifold} encoder leads with 86.2\% accuracy and a macro-F1 of 0.84, followed closely by Persistent Homology at 85.8\% (macro-F1 0.82). The separation among GNN variants is modest (4.6 percentage points) suggesting that under in-distribution conditions, the specific choice of structural encoding matters less than the fundamental strength of the multimodal ESM+GAT backbone.

\begin{table*}[t]
\centering
\caption{\textbf{In-Distribution Performance:} Random split (80/20, stratified, seed=42). Results report overall accuracy, per-class F1 scores, and macro F1 based on the optimal checkpoint selected within a maximum 50-epoch budget using early stopping. Best results per column are bolded.}
\label{tab:random_split}
\begin{tabular}{llcccccc}
\toprule
\textbf{Rank} & \textbf{Model} & \textbf{Test Acc} & \textbf{Ca F1} & \textbf{K F1} & \textbf{Na F1} & \textbf{Macro F1} & \textbf{Best Epoch} \\
\midrule
1 & \textbf{(M1) Stat. Manifold} & \textbf{86.2\%} & \textbf{0.73} & \textbf{0.93} & 0.86 & \textbf{0.84} & 9/50 \\
2 & (M2) PH-GNN & 85.8\% & 0.69 & 0.90 & \textbf{0.88} & 0.82 & 15/50 \\
3 & (M3) Spectral & 84.0\% & 0.68 & 0.90 & 0.85 & 0.81 & 11/50 \\
4 & Baseline (ESM+GAT) & 82.9\% & 0.68 & \textbf{0.92} & 0.83 & 0.81 & 11/50 \\
5 & (M4) Riemannian & 82.9\% & 0.57 & 0.90 & 0.84 & 0.77 & 4/50 \\
6 & (M5) Invariant SE(3) & 81.1\% & 0.65 & 0.91 & 0.81 & 0.79 & 9/50 \\
\bottomrule
\end{tabular}
\end{table*}

\subsection{Out-of-Distribution: Taxa Leave-One-Out}
Table \ref{tab:ood} reports OOD accuracy. The contrast with the random-split results is dramatic. Five of the six GNN architectures cluster closely between 54.1\% and 55.4\% (a difference of only 1.3 percentage points). In stark contrast, \textbf{Persistent Homology} stands apart at 63.4\%, commanding an 8 percentage point margin over all other representations. Furthermore, the ESM+RF baseline, which was historically competitive under random splits, collapses to $\sim$30\% mean OOD accuracy, including a catastrophic 6.3\% on scorpion proteins (a statistically robust fold containing 646 proteins). 

\textbf{Per-class Analysis.} Homology's advantage is consistent across most folds. It achieves the highest accuracy on Snake (75.3\%), Scorpion (74.1\%), and Sea Anemone (69.6\%), and remains competitive on Cone Snail (51.2\%); the Spider fold, discussed below, is the one exception. Crucially, homology achieves a Ca F1 of 0.81 on the Snake fold, the highest single class-level score across the entire OOD evaluation, suggesting the topology representation is genuinely learning transferable functional features rather than relying on properties specific to one fold. (Note: Sodium F1 approaches zero for the Snake fold across all models due to data sparsity (fewer than 10 Na proteins exist in the Snake test set) reflecting a dataset limitation rather than a model failure.)

\textbf{The generalization penalty.} We can quantify the phylogenetic trap by examining the gap between random-split and OOD performance. The Riemannian model achieves 82.9\% on the random split but only 54.1\% OOD, yielding a 28.8 percentage point drop. For Homology, the gap is significantly narrower at 22.4 percentage points (85.8\% $\rightarrow$ 63.4\%). This smaller OOD penalty indicates a highly robust representation that relies far less on taxon-specific features.

\begin{table*}[t]
\centering
\setlength{\tabcolsep}{4pt}
\caption{\textbf{Out-of-Distribution Performance:} Taxa leave-one-out cross-validation accuracy (\%) per taxa, mean among all taxa and the standard deviation for all the models. Last fold for baseline 7 was out of memory. McNemar exact test vs.\ PH-GNN on paired per-protein predictions pooled
  across folds ($N=1{,}373$), Holm-corrected.}
\label{tab:ood}
\begin{tabular}{llcccccccc}
\toprule
\textbf{Rank} & \textbf{Model} & \textbf{Snake} & \textbf{Cone Snail} & \textbf{Spider} & \textbf{Scorpion} & \textbf{Sea Anemone} & \textbf{Mean} & \textbf{Std} &  $p_{\text{holm}}$\\
\midrule
1 & \textbf{(M2) PH-GNN} & \textbf{75.3\%} & 51.2\% & \textbf{46.7\%} & \textbf{74.1\%} & 69.6\% & \textbf{63.4\%} & 12.0\% & ---\\
2 & (M5) Invariant SE(3) & 49.4\% & \textbf{52.0\%} & 41.4\% & 61.1\% & \textbf{73.3\%} & 55.4\% & 10.9\% & $6.3\times10^{-14}$ \\
3 & (M1) Stat. Manifold & 46.8\% & 46.5\% & 43.1\% & \textbf{70.0\%} & 69.6\% & 55.2\% & 12.0\% & $2.7\times10^{-5}$\\
4 & Baseline (ESM+GAT) & 45.5\% & \textbf{52.0\%} & 43.1\% & 66.7\% & 68.9\% & 55.2\% & 10.7\% & $4.8\times10^{-17}$\\
5 & (M3) Spectral & 48.1\% & 48.8\% & 41.2\% & 62.8\% & 70.8\% & 54.3\% & 10.8\% & $2.6\times10^{-9}$\\
6 & (M4) Riemannian & 45.5\% & 52.8\% & 41.4\% & 61.9\% & 68.9\% & 54.1\% & 10.2\% & $7.9\times10^{-11}$ \\
\midrule
7 & ESM+RF (Seq. Only) & 45.1\% & 45.1\% & 24.0\% & 6.3\% & N/A & $\sim$30\% & - & -\\
\bottomrule
\end{tabular}
\end{table*}

\textbf{The Spider fold limitation.} The Spider fold is the hardest case for all architectures, with persistent homology achieving 46.7\% accuracy versus a 44.8\% Sodium-majority baseline. While topology generalizes substantially better than competing inductive biases on four of the five folds, even topological invariants cannot fully escape the phylogenetic trap when the held-out taxon contains exceptional fold diversity. Spider compounds are dominated by the inhibitor cystine knot (ICK) scaffold, but ICK loop topology varies substantially across spider families, producing a held-out distribution whose multi-scale persistence signatures are themselves heterogeneous. This represents a remaining limitation rather than a contradiction of our central thesis: persistent homology reduces but does not eliminate cross-species generalization gaps, and the magnitude of the residual gap depends on the structural diversity of the held-out taxon. Closing this gap likely requires explicit modeling of higher-order disulfide connectivity, which we identify as the most promising direction for future work.
\subsection{Taxa-Class entanglement in embedding space}
\begin{figure*}[t]
    \centering
    
    \includegraphics[width=0.79\textwidth]{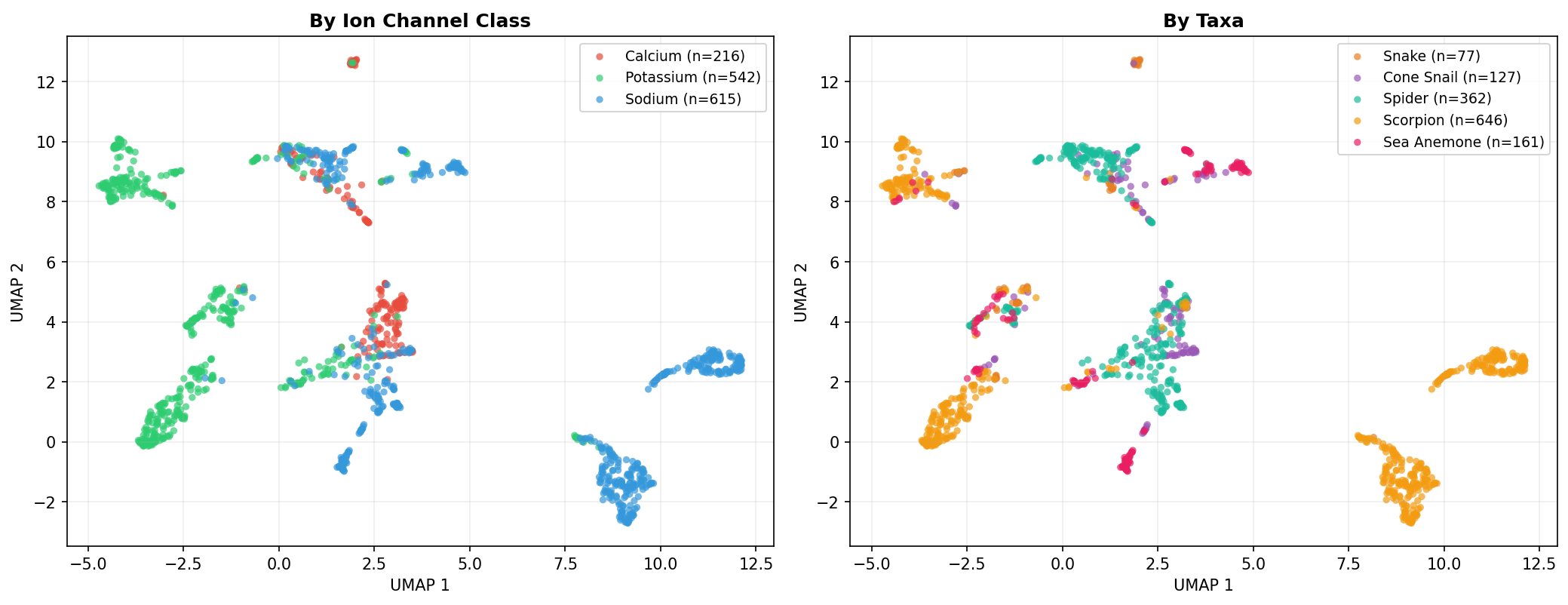} 
    \caption{\textbf{Visualization of the phylogenetic trap.} UMAP projection of the learned 512-dimensional concatenated structure and sequence embeddings (prior to fusion) from the baseline multimodal GNN. \textit{Left:} Colored by ion channel target. \textit{Right:} Colored by organism taxon. The strong clustering by taxon rather than target class visually demonstrates the model's reliance on evolutionary lineage, explaining the catastrophic collapse during out-of-distribution evaluation.}
    \label{fig:umap_entanglement}
\end{figure*}

To explicitly understand why random split accuracy is inflated, we apply UMAP \cite{McInnes2018} to the learned 512-dimensional concatenated structure and sequence embeddings (prior to fusion) (Figure \ref{fig:umap_entanglement}). 

The resulting 2D projections reveal that proteins cluster predominantly by organism taxon: Scorpion proteins form a dense, well-separated cluster; Snake proteins form another; Cone Snail proteins yet another, regardless of their true ion channel target class. While within-taxon class boundaries are learnable (explaining the high random-split accuracy), between-taxon transfer is not automatic. This directly visualizes the OOD collapse: a model trained without Scorpion proteins has no embedded representation for the Scorpion region of the feature space, forcing it to fall back on majority-class predictions. Persistent homology is fundamentally less susceptible to this entanglement because the Alpha complex persistence diagram of $C\alpha$ coordinates is invariant to amino acid identity. It responds exclusively to the geometric arrangement of backbone atoms, which is shaped by the target channel geometry rather than the organism's evolutionary history.
\subsection{Statistical significance} We compare PH-GNN against each variant using McNemar's exact test \cite{mcnemar1947} on paired per-protein predictions pooled across the five held-out taxa (N = 1,373), with Holm–Bonferroni correction \cite{holm1979}. PH-GNN significantly outperforms all five variants (all $p < 3\times10^{-5}$), with pooled accuracy gains of +4.95 to +9.40 points; the gain over the baseline is +9.40 points (bootstrap 95\% CI [7.28, 11.51]). Per fold tests localize the improvement to Snake (+23.4, $p=0.002$) and Scorpion (+15.8, $p<10^{-4}$), with no significant difference on Cone Snail, Spider, or Sea Anemone — topology helps most where the baseline degenerates to single-class prediction. A fold-level paired $t$-test is underpowered at $n=5$ (all $p>0.14$ despite $d_z=0.67$–$0.81$). Significance is computed over test proteins.
\section{Discussion}

\subsection{Why Topology Generalizes Cross-Species}
The fundamental reason persistent homology generalizes where other methods fail lies in the biophysics of ion selectivity. Ion selectivity is mechanistically determined by the rigid geometry of the target channel's selectivity filter: potassium channels require a specific spacing of backbone carbonyl oxygens ($\sim$3 \AA) to coordinate $K^{+}$ ions via partial dehydration \cite{Doyle1998}, while calcium channels require a ring of glutamate residues creating a specific electrostatic well \cite{Yang1993}. Compounds targeting these channels must converge on complementary geometric solutions: specific arrangements of loops and hairpins that match the pore's topological profile. 

Persistent homology, computed over an alpha complex filtration up to a circumradius of $20$~\AA, captures the formation and destruction of these cavities and loops.
It explicitly encodes the multi-scale topological structure of the interaction surface. Crucially, this structure is a direct consequence of the target channel's deeply conserved geometry, completely bypassing the compound's evolutionary lineage. In contrast, other representations fall into the phylogenetic trap. Spectral features (Laplacian eigenvectors) encode the global shape of the entire protein graph, which is highly scaffold-dependent and therefore taxon-correlated. $SE(3)$-invariant distance and angle features are rotation-free but remain tied to local species-specific geometry. Sequence-derived features (ESM-2, whether used in RF or Fisher-geometry form) are inherently evolutionary; they encode which mutations are tolerated, which depends heavily on the organism's codon usage, expression context, and phylogenetic constraints.

\subsection{Explainable AI: Visualizing Topological Attention}
\begin{figure}[t]
    \centering
    
    \includegraphics[width=0.95\columnwidth]{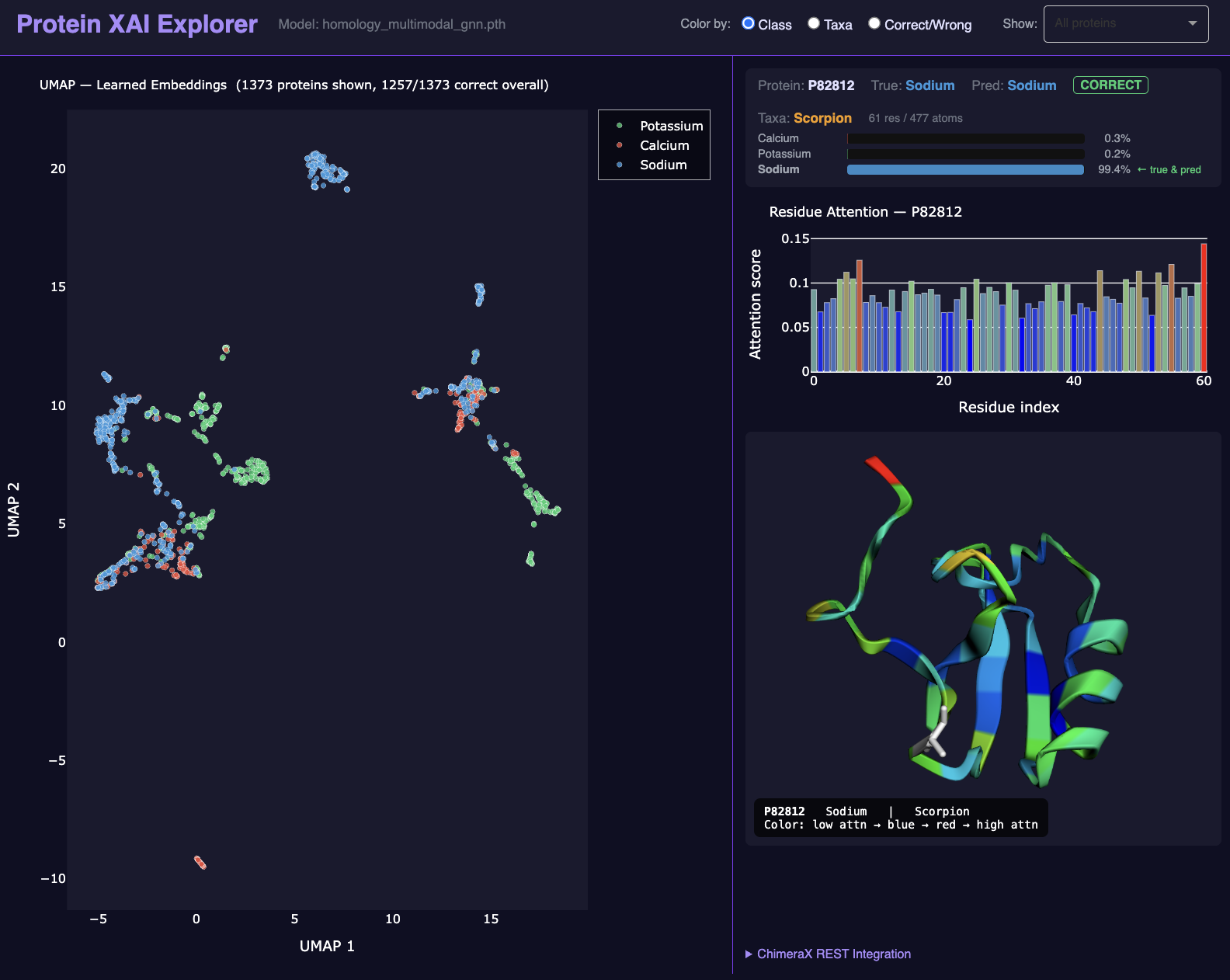}
    \caption{\textbf{Residue-Level Explainable AI (XAI).} Our interactive visualization framework maps attention weights directly onto the ColabFold-predicted 3D structure. The residue attention bar chart (top right) and structural heatmap (bottom right) reveal exactly which topological loops and explicit disulfide bonds drive the prediction, bridging the gap between abstract mathematical embeddings and actionable biophysics.}
    \label{fig:xai_dash}
\end{figure}

To validate that our multimodal architecture successfully captures these functional interaction surfaces rather than arbitrary dataset artifacts, we developed an interactive Explainable AI (XAI) framework (Figure \ref{fig:xai_dash}). 
By projecting the GAT attention weights back onto the ColabFold-predicted 3D structures, our residue-level dashboard reveals the physical mechanisms driving the predictions. For successful cross-species classifications, the attention heads consistently highlight the precise structural motifs (such as the cysteine-stabilized functional hairpins) that physically occlude the target ion channel pore. Furthermore, the visualization confirms that explicitly weighting disulfide bonds successfully forces the network to attend to the rigid thermodynamic scaffolds that preserve these topological loops, confirming our theoretical intuition.

\subsection{The Evaluation Trap and Practical Implications}
The near-equivalence of architectures under the random split (81 to 86\%, Table \ref{tab:random_split}), followed by the dramatic divergence under OOD evaluation (54 to 63\%, Table \ref{tab:ood}), illustrates a critical vulnerability in biological machine learning: standard random splits permit homologous sequence leakage that artificially inflates performance \cite{Koh2021}. The 22 to 29 percentage point OOD penalty observed in our baseline models is unusually large, reflecting the extreme taxa diversity within venomics. Based on these findings, we strongly recommend taxa-stratified OOD evaluation as the minimum benchmark standard for any multi-species protein classification task. Ultimately, the choice of structural representation must be dictated by the deployment scenario. When a researcher applies a trained classifier to a newly sequenced venom proteome from an unstudied species (e.g., field venomics or biodiversity annotation), the deployment perfectly mirrors our OOD protocol. Under those discovery conditions, homology encoding provides an 8 percentage point advantage. Conversely, for high-throughput laboratory screening over organisms already well-represented in the training distribution, the performance differences shrink, and highly optimized in-distribution models like the Statistical Manifold may be utilized for maximum precision.
\section{Limitations and Future Work}

While our framework successfully bridges the cross-species generalization gap, future work will address channel promiscuity via multi-label classification and explicitly model Inhibitor Cystine Knot entanglements using knot theory. Additionally, we aim to capture the true temporal dynamics of peptide-ion channel binding by applying cohomology to structural ensembles and leveraging 3D structural world models.
\section{Conclusion}

We present a \textbf{novel multimodal representation learning framework} for peptide-ion channel classification. By fusing ESM-2 embeddings with structure-aware GATs and explicit disulfide bond attention, we demonstrate that the choice of evaluation protocol fundamentally dictates which structural inductive bias succeeds. Through a \textbf{systematic evaluation of geometric priors} (comparing statistical manifolds, persistent homology, spectral graph theory, Riemannian manifolds, and SE(3) invariance), we show that these mathematical abstractions behave similarly under standard random splits. Architectures differ by at most 5 percentage points, with the Statistical Manifold successfully exploiting evolutionary lineage as a confounder to memorize sequence distributions. However, when evaluated under a strict taxonomic shift (taxa leave-one-out OOD), a catastrophic phylogenetic trap is revealed where sequence-only baselines and standard geometric graph methods collapse. \textbf{Escaping this OOD trap}, our persistent homology framework (PH-GNN) stands 8 percentage points above all alternatives, achieving a leading 63.4\% mean cross-species accuracy and proving that topology generalizes where local geometry fails. The key mechanism driving this success is topological invariance. Cavity persistence diagrams encode the multi-scale geometry of the compound interaction surface, enabling the network to perform higher-level reasoning about complex structural relationships. This representation converges across evolutionary lineages because it is constrained by the deeply conserved geometry of the target ion channel pores.
Ultimately, we reveal a failure mode in biological ML evaluation and identify persistent homology as a superior structural inductive bias for domain generalization.

\section*{Acknowledgments}
This work used Bridges-2 at the Pittsburgh Supercomputing Center through allocation bio230056p from the Advanced Cyberinfrastructure Coordination Ecosystem: Services \& Support (ACCESS) program, which is supported by U.S. National Science Foundation grants \#2138259, \#2138286, \#2138307, \#2137603, and \#2138296. This research and its open accessibility were funded by the National Institutes of Health Pioneer Award (grant 5DP1AT012812 to MH) and MBL Whitman Fellowship to MH.

\bibliography{references}

\end{document}